\documentclass{article}
\usepackage{spconf,amsmath,graphicx}
\usepackage{epstopdf}
\usepackage{times}
\usepackage{amssymb}
\usepackage{enumitem}
\usepackage{url}
\usepackage{subcaption}

\title{Partial Face Detection for Continuous Authentication}

\name{Upal Mahbub$^{\star}$ \quad Vishal M. Patel$^{\dagger}$  \quad Deepak Chandra$^{\mathsection}$  \quad Brandon Barbello$^{\mathsection}$  \quad Rama Chellappa$^{\star}$}
\address{$^{\star}$Department of Electrical and Computer Engineering and the Center for Automation Research, \\UMIACS, University of Maryland, College Park, MD 20742\\$^{\dagger}$Rutgers, The State University of New Jersey, 723 CoRE, 94 Brett Rd, Piscataway, NJ 08854\\$^{\mathsection}$Google Inc., 1600 Amphitheatre Parkway,  Mountain View, CA 94043 \\
{\tt\small umahbub@umiacs.umd.edu, vishal.m.patel@rutgers.edu,}\\
{\tt \small dchandra@google.com, bbarbello@google.com, rama@umiacs.umd.edu} \thanks{This work was done in partnership with and supported by Google Advanced Technology and Projects (ATAP), a Skunk Works-inspired team chartered to deliver breakthrough innovations with end-to-end product development based on cutting edge research and a cooperative agreement FA8750-13-2-0279 from DARPA.} }

\begin{document}
\ninept
\maketitle
\begin{abstract}
In this paper, a part-based technique for real time detection of users' faces on mobile devices is proposed. This method is specifically designed for detecting partially cropped and occluded faces captured using a smartphone's front-facing camera for continuous authentication. The key idea is to detect facial segments in the frame and cluster the results to obtain the region which is most likely to contain a face. Extensive experimentation on a mobile dataset of 50 users shows that our method performs better than many state-of-the-art face detection methods in terms of accuracy and processing speed.
\end{abstract}
\section{Introduction}
Face detection, being one of the most fundamental problems in computer vision, plays an important role in the performance of various face-based applications ranging from face identification and verification to clustering, tagging and retrieval \cite{FaceevaluationStanLi}. There has been substantial progress in the development of efficient face detection techniques and many such techniques are now available in commercial products such as digital cameras, smartphones and social networking websites \cite{fddbTech}. Both in academia and industry, the current trend of developing new face detectors is primarily centered around detecting faces in unconstrained environments with huge variations of pose and illumination. This has led to the development of a number of recent face detectors that try to outperform each other on the Labeled Faces in the Wild (LFW) \cite{LFWTech}, Annotated Facial Landmarks in the Wild (AFLW) \cite{AFLWDataset} and other such datasets. Another focus of current face detection research is in developing fast face detectors for real time implementation on devices such as smartphones, tablets etc.

In today's world, mobile devices are being used not only for verbal communication but also for accessing bank accounts and performing transactions, managing user profiles, accessing e-mail accounts, etc. With increasing usage, there is a growing concern about ensuring the security of users' personal information on these devices. Going beyond passwords and fingerprints, concepts have emerged for actively verifying users by analyzing faces from the front-facing camera, the swipe and keystroke patterns from the touchscreen and motions patterns from inertial sensors, among others \cite{umd_Dataset}, \cite{ContAuth_AJain}, \cite{Mobio_2012}. Face-based authentication systems rely on accurate detection of faces in their first step and successful verification in the next. Several research works have recently been published on face-based continuous user authentication techniques for smartphones, using representation-based or attribute-based approaches \cite{AA_Fathy}, \cite{Mobio_2012}, \cite{AA_Samangouei}. However, in all of these methods the faces and landmarks are assumed to be available by performing face detection within milliseconds.

\begin{figure}[t]
\centering
\includegraphics[width=0.25\textwidth]{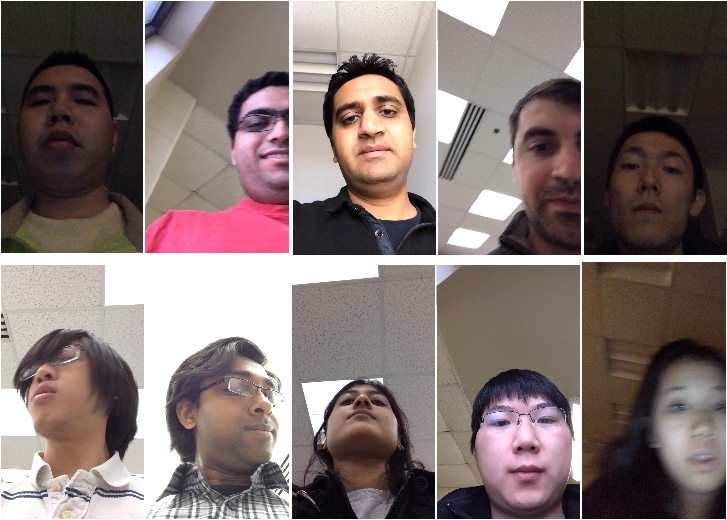}
\vskip -5pt
\caption{Sample frames from the AA-01-FD mobile face dataset where one can clearly see the presence of partial faces.}
\label{UMDDataSample}
\vskip -10pt
\end{figure}
Viola and Jones's \cite{VJFull} ground-breaking method of face detection popularized Adaboost cascade structures and simple feature extraction techniques for realtime face detection. Hadid \emph{et al.} \cite{AA_Hadid} evaluated the feasibility of face and eye detection on cell phones using Adaboost cascade classifiers with Haar-like and local binary pattern (LBP) features, as well as a skin color based face detector. On a  Nokia N90 mobile phone that has an ARM$9$  $220$  MHz processor and  a  built-in memory of $31$ MB, their work reported that the Haar+Adaboost method can detect faces in $0.5$ seconds from $320\times 240$ pixel images. This approach, however, is not effective when wide variations in pose and illumination are present. Some variations of this method are available in the literature such as \cite{RotationInvMultiview_Huang}, \cite{Multiview_heyden} but most of them are computationally very expensive. Zhu \emph{et al.} \cite{Ramanan:2012:FDP:2354409.2355119} proposed a method that uses a deformable part model where a face is defined as a collection of parts at each facial landmark, and the parts are then trained side-by-side with the face using a spring-like constraint. The method uses a mixture of tree-structured models resilient to viewpoint changes.  Mar$\acute{i}$n-Jim$\acute{e}$nez \emph{et al.} \cite{LAEOdataset} also followed a similar approach for `Looking At Each Other' head detection. Whereas, Shen \emph{et al.} \cite{Shen_ExamplarFD} proposed an exemplar-based face detector that exploits advanced image retrieval techniques to avoid an expensive sliding window search. Apart from academic research, many face detectors such as Google Picasa, PittPatt and Face++ have been developed commercially as face detection has gotten more and more attention over time. However, most of the state-of-the-art techniques are not designed to detect partially visible or cropped faces (see Figure~\ref{UMDDataSample}). Hence, their recall rate is usually very low when operating at high precision.  Furthermore, many of these methods require long computation times because of algorithmic complexity and are thus not suitable for real time implementation.

In this paper, we introduce a face detection scheme for detecting faces from images captured by the front-facing cameras of mobile devices used for continuous authentication.  This paper makes the following contributions:   (1) A method for realtime face detection on a smartphone is proposed based on facial segments that can detect cropped and partial faces of users. (2)  A simple yet effective technique of clustering the facial segments is proposed. (3) Through extensive experimentation, it is shown that the proposed method can achieve superior performance over some of the state-of-the-art methods in terms of accuracy and processing time.

In the next section the proposed method is described in detail followed by a brief summary of the dataset. Then, the  experimental setup, and results are explained. Finally, a summary of the work with future directions is provided in the section on conclusion.
\section{Facial Segment based Face Detector (FSFD)}\label{method}
\begin{figure}[t]
\centering
\includegraphics[width=0.38\textwidth]{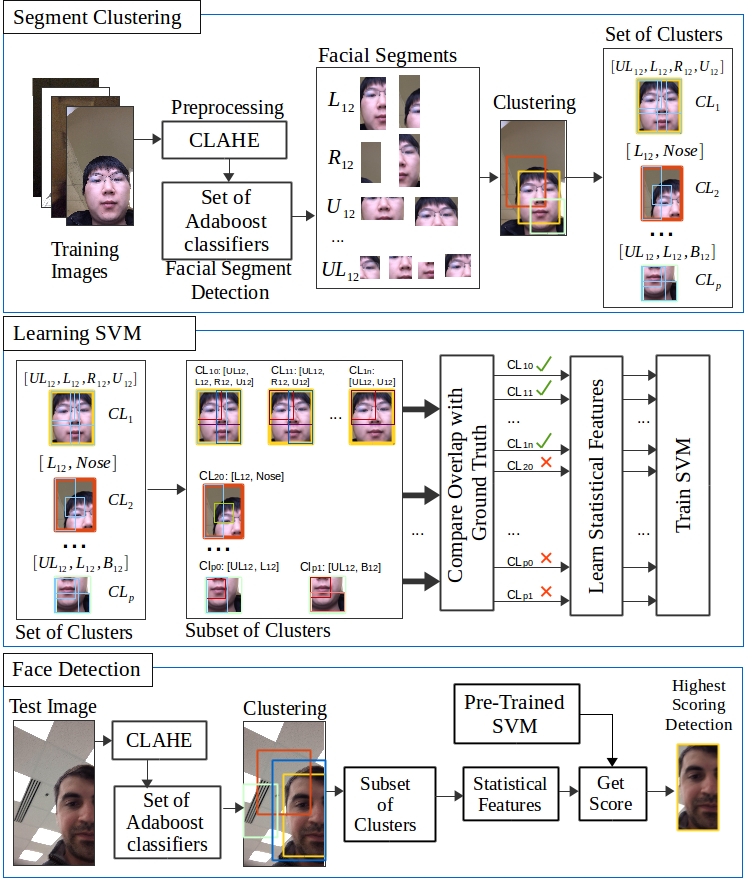}
\vskip -8pt
\caption{The block diagram of the overall system.}
\label{SysteDiagramSVM}
\vskip -10pt
\end{figure}
The system block diagram for the proposed face detector is shown in Fig. \ref{SysteDiagramSVM}. The process is divided into three phases. In the segment clustering phase, facial segments are logically clustered to estimate face regions from the training images. In the Support Vector Machine (SVM) learning phase, a linear SVM is trained with statistical features obtained from face proposals that are derived from the estimated faces. Finally, in the face detection phase, statistical features are obtained in a similar manner for each face proposal from test images and confidence scores are obtained for each proposal using the SVM classifier. The proposal with the highest confidence score higher than a certain threshold $\theta$ is considered to be a face. Because of its simple architecture, the method can be implemented in realtime. In addition, it is fairly accurate and provides a good recall rate since it can efficiently detect partially visible faces. Thus, the FSFD method is suitable for real time face verification tasks on mobile devices. The three main steps of the detector are discussed in detail in the following subsections.
\subsection{Segment Clustering}
\begin{figure}[t]
\centering
\includegraphics[width=0.27\textwidth]{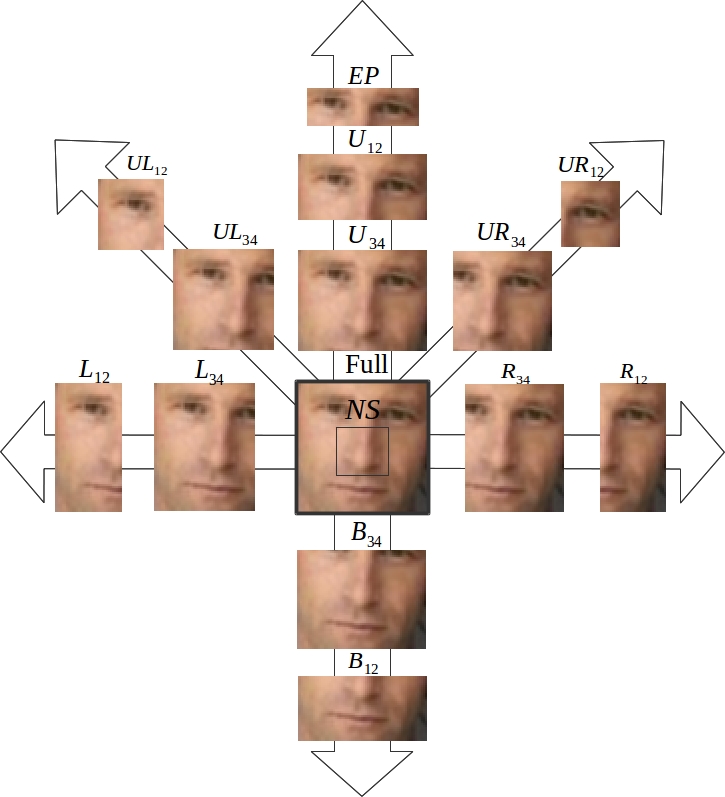}
\vskip -8pt
\caption{All of the $14$ facial segments.}
\label{FaceCrop}
\vskip -10pt
\end{figure}

As shown in Fig. \ref{SysteDiagramSVM}, in the segment clustering phase, contrast limited adaptive histogram equalization (CLAHE) is performed first on each of the training images to reduce the impact of illumination on facial segment detectors. Each training image is then passed through a set of facial segment detectors. A total of $14$ facial segment detectors were trained using facials segments obtained from the cropped and aligned Labeled Faces in the Wild (LFW) dataset \cite{LFWTech}, \cite{LFW_Sanderson} and negative examples were produced from the Pascal VOC dataset \cite{PascalVOC}. These detectors are basically adaboost cascade classifiers trained using a local binary pattern (LBP) representation of the images for better feature representation and faster training \cite{LBPVJ}, \cite{opencv_library}. The facial segments for which the classifiers were trained are Eye-pair ($EP$), Upper-left-half of face ($UL_{12}$), Upper-half of face ($U_{12}$), Upper-right-half of face ($UR_{12}$), Upper-left-three-fourth of face ($UL_{34}$), Upper-three-fourth of face ($U_{34}$), Upper-right-three-fourth of face ($UR_{34}$), Left-half of face ($L_{12}$), Left-three-fourth of face ($L_{34}$), Nose ($NS$), Right-three-fourth of face ($R_{34}$), Right-half of face ($R_{12}$), Bottom-three-fourth of face ($B_{34}$) and Bottom-half of face ($B_{12}$). An example of all the $14$ facial segments of a full face from the LFW dataset is shown in Fig. \ref{FaceCrop}.
Each detector may return one or more facial segments in a frame. For example, the segment clustering phase of Fig. \ref{SysteDiagramSVM},  $L_{12}$ returned two detection results while $UL_{12}$ returned four detection results. For each facial segment, the bounding box of the full face is estimated according to the ideal position of that segment relative to the whole face. For example, if the top left and bottom right corners of the bounding box obtained for segment $L_{12}$ are ($x_{1}^{L12}, y_{1}^{L12}$) and ($x_{2}^{L12}, y_{2}^{L12}$), respectively, then those for the estimated full face are ($x_{1}^{L12}, y_{1}^{L12}$) and ($min(w_{img}, x_{2}^{L12}+(x_{2}^{L12}-x_{1}^{L12})), y_{2}^{L12}$), where $w_{img}$ is the width of the image. The estimated face center from this segment is $(x_{2}^{L12}, y_{1}^{L12}+(y_{2}^{L12}-y_{1}^{L12})/2)$. For each estimated face center $p$, a cluster of segments $CL_{p}$ that depicts the full face is formed where (a) the number of segments in that cluster is above a threshold $c$, and (b) the other segments of that cluster have estimated face centers within a certain radius $r$ pixels from the center.
\subsection{Learning SVM}
In the learning phase, the first $\zeta$ subsets of the total set of facial segments from each cluster are regarded as proposed faces. Assuming that $m$ facial segments satisfy the condition for clustering around the $k$-th segment, there are $m+1$ segments in that cluster where $m+1 \geq c$. For example, if the $4$ segments $U_{12}$, $B_{12}$, $L_{34}$ and $UR_{12}$ form a cluster around $NS$ and $c=2$, then the viable subsets are $\allowbreak[[NS, U_{12}],\allowbreak [NS, B_{12}],\allowbreak \hdots,\allowbreak,\allowbreak [NS,\allowbreak U_{12}, \allowbreak B_{12},\allowbreak L_{34},\allowbreak UR_{12}]]$. The total number of subset here is $\sum_{j=1}^{4}{{4}\choose{j}}= 15$ including the complete set. Keeping the $k$-th segment fixed, $\zeta$ random subsets are considered for face proposals. In this example, $\zeta$ varies from $1$ to $14$. For $m+1$ segments, the number of subsets is in the order of $2^{m+1}$, which introduces huge computation and memory requirements. Hence, the number of subset is limited to $\zeta << 2^{m+1}$.

The bounding box of the face proposal is the smallest bounding box that encapsulates all the estimated faces from all the facial segments in that proposal. Intuitively, the greater the number of facial segments with better detection accuracy in a proposal, the higher the probability of that proposal being a face. Further, experimentally, it is found that some particular sets of facial segments are more likely to return a face than others, and some sets of segments provide more accurate bounding boxes with greater consistency than the others. The amount of overlap is expressed as the ratio of intersection areas to joined area $I(d_{i}, a_{j})$, between the detected region $d_{i}$ and annotated regions $a_{j}$, and is expressed as
\vskip -8pt
\begin{equation}
I(d_{i}, a_{j})=\frac{area(d_{i})\cap area(a_{j})}{area(d_{i})\cup area(a_{j})}.
\end{equation}
If this ratio is above a certain threshold $\delta$, then the detection result is considered to be correct.

A linear SVM classifier is trained on the proposed faces using the following statistical features.
\begin{enumerate}
\item Probability of a proposal $S_{set}$ being a face $Pr^{T}(S_{set})$ and not being a face $Pr^{F}(S_{set})$ are defined as
\vskip -15pt
\begin{eqnarray}
Pr^{T}(S_{set})&=&\frac{\sum{|S_{set}\in P_{F}^{T}|}}{|P_{F}^{T}|}\\
Pr^{F}(S_{set})&=&\frac{\sum{|S_{set}\in P_{F}^{F}|}}{|P_{F}^{F}|}
\end{eqnarray}
where, $P_{F}^{T}=\{S|I(S, S_{GT})\geq \delta; S\in P_{Tr}^{F}\}$, and $P_{F}^{F}=\{S|I(S, S_{GT})<\delta; S\in P_{Tr}^{F}\}$. Here, $P_{Tr}^{F}$ denotes the set of clusters of facial segments that the detector proposed as faces.
\item For each facial segment in a set $P^{i}_{Set}$ where $i\in S_{set}$, the probabilities of $P^{i}_{Set}$ being in a cluster depicting the true face $Pr^{T}(P_{set}^{i})$ or a non-face $Pr^{F}(P_{set}^{i})$ are calculated as
\vskip -15pt
\begin{eqnarray}
Pr^{T}(P_{set}^{i})=\frac{\sum{|P^i_{set}\in S_{set}; S_{set}\in P_{F}^{T}|} }{|P_{F}^{T}|}\\
Pr^{F}(P_{set}^{i})=\frac{\sum{|P^i_{set}\in S_{set}; S_{set}\in P_{F}^{F}|}}{|P_{F}^{F}|}.
\end{eqnarray}
Experimentally, it is found that the nose detector is the most accurate of all, while $B_{12}$ is the least accurate.
\end{enumerate}
If $n$ segments are considered then the feature size is $2n+2$ for each proposal. There are $2n$ values corresponding to the face and non-face probabilities of each of the $n$ segment and the rest $2$ values are the probabilities of the cluster being and not-being a face. Among the $2n$ values, only those corresponding to the segments present in the proposal are non-zero.
\subsection{Face Detection}
For each pre-processed test image, the proposed faces are obtained in a similar manner as the training faces. Thus, there are $\zeta$ face proposals from each face and feature vectors of size $2n+2$ for each proposal. The SVM classifier returns a confidence score for each proposal. The proposal that has the highest confidence score above a threshold is chosen as the detected face.
\section{Dataset} \label{Dataset}
The performance of the proposed face detector is evaluated on the Active Authentication Dataset (AA-01) \cite{umd_Dataset}, \cite{AA_Samangouei}. The AA-01 Dataset contains the front-facing camera face video for $50$ iphone users ($43$ male, $7$ female) with three different ambient lighting conditions: well-lit, dimly-lit, and natural daylight. In each session, the users performed $5$ different tasks: enrollment, scroll test, picture count, document reading and picture drag. In order to evaluate the face detector, face bounding boxes were annotated in a total of $8036$ frames of the $50$ users. This dataset, denoted as AA-01-FD, contains $1607$ frames without faces and $6429$ frames with faces. For training, $20\%$ of these frames are randomly picked and the rest are used as test data. Some sample face images from the AA-01-FD Dataset are shown in Fig. \ref{UMDDataSample}.
\section{Experimental Results} \label{Result}
The experimental results are presented in this section, demonstrating the effectiveness of the FSFD method over other state-of-the-art face detectors. In particular, experimental results on the AA-01-FD dataset are compared with a. the Viola-Jones face detector (VJ) \cite{VJFull}, b. the deformable part-based model face detector (DPM) \cite{Ramanan:2012:FDP:2354409.2355119}, and c. the Looking At Each Other (LAEO) head detector \cite{LAEOdataset}.
\subsection{Evaulation Metrics and Experimental Setup}
The results are evaluated in terms of F1-Score, true positive rate (TPR) at $1\%$ false positive rate (FPR), recall at $99\%$ precision, and processing time. The F1-Score, which is the harmonic mean of precision and recall, is defined as
\vskip -8pt
\begin{equation}
F1=\frac{2TP}{2TP+FP+FN}
\end{equation}
where, $TP$, $FP$ and $FN$ are the numbers of true positive,  false positive and false negative detections, respectively. Higher F1-Score employs better overall precision and recall. For the active authentication problem, the goal is to achieve the best recall at a very high precision. Hence, the value of recall achieved by a detector at $99\%$ precision is considered another evaluation metric. Finally, for detectors that return a confidence score rather than a binary face/non-face decision, the TPR at $1\%$ FPR is used as a metric for evaluation. Here, also, higher values of TPR represent a better detector.

\begin{figure}[t]
\centering
\includegraphics[width=0.43\textwidth]{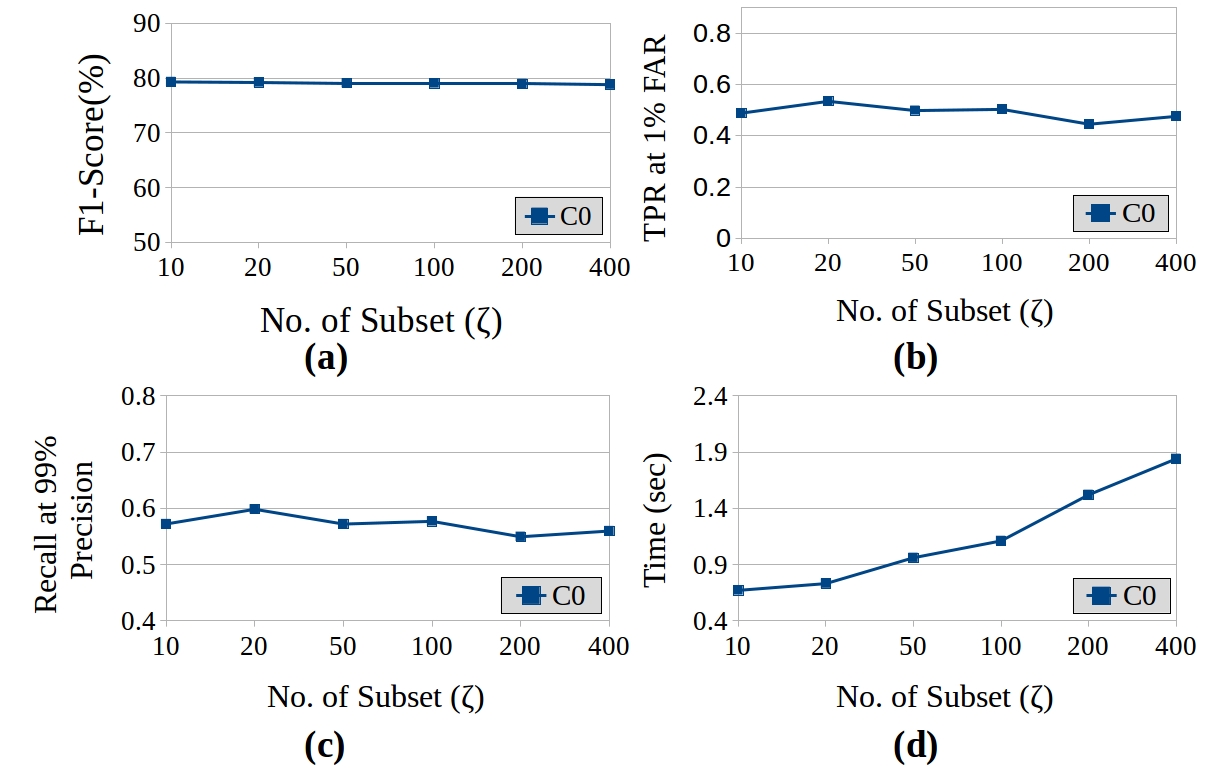}
\caption{Performance of $C0$ for different values of $\zeta$ - (a) F1-Score vs. $\zeta$, (b) TPR at $1\%$ FAR vs. $\zeta$, (c) Recall at $99\%$ Precision vs. $\zeta$, and (d) Time vs. $\zeta$.}
\label{INVariation}
\end{figure}

Based on the relative importance of each segment, several combinations of segments are considered to determine the optimum number of segments to produce the best result. The basic combination of all $14$ segments is labeled as $C0$ and the best performing combination is labeled $Cbest$. $Cbest$ consists of $9$ segments - $NS$, $EP$, $UL_{34}$, $UR_{34}$, $U_{12}$, $L_{34}$, $UL_{12}$, $R_{12}$, $L_{12}$.

All the experiments in this paper were performed on an Intel Xeon(R) CPU E5506 operating at $2.13$GHz with $5.8$GB memory and a $64$ bit UBUNTU $14.04$ LTS operating system. No multi-threading was done to parallelize the operations of different facial segment detectors, hence, speed up is achievable \cite{Pulli:2012:RCV:2184319.2184337}. The images, initially $1280\times 720$ pixels, are downsampled by $4$. It is experimentally verified that for all the methods this downsampling ensures best performance in quickest time. The minimum face size is set to be $64\times 64$ pixels. For clustering facial segments, the value of $c$ is set to two and $r$ is considered to be one-sixth of the half-diagonal of the estimated face of the center segment. A lower value of $r$ may improve the precision while possibly decreasing the recall.

In order to determine the optimum value of $\zeta$, the evaluation metrics and average processing time per image are plotted in Fig. \ref{INVariation} with increasing $\zeta$ for combination $C0$ on AA-01-FD. It can be inferred that the performance of the system does not change significantly, while the time consumption per image rapidly increases with $\zeta$. Hence, a value of $\zeta=20$ is chosen to ensure the best performance at the smallest time.
\subsection{Results and Comparison}\label{Results}
\begin{figure}[t]
\centering
\includegraphics[width=0.4\textwidth]{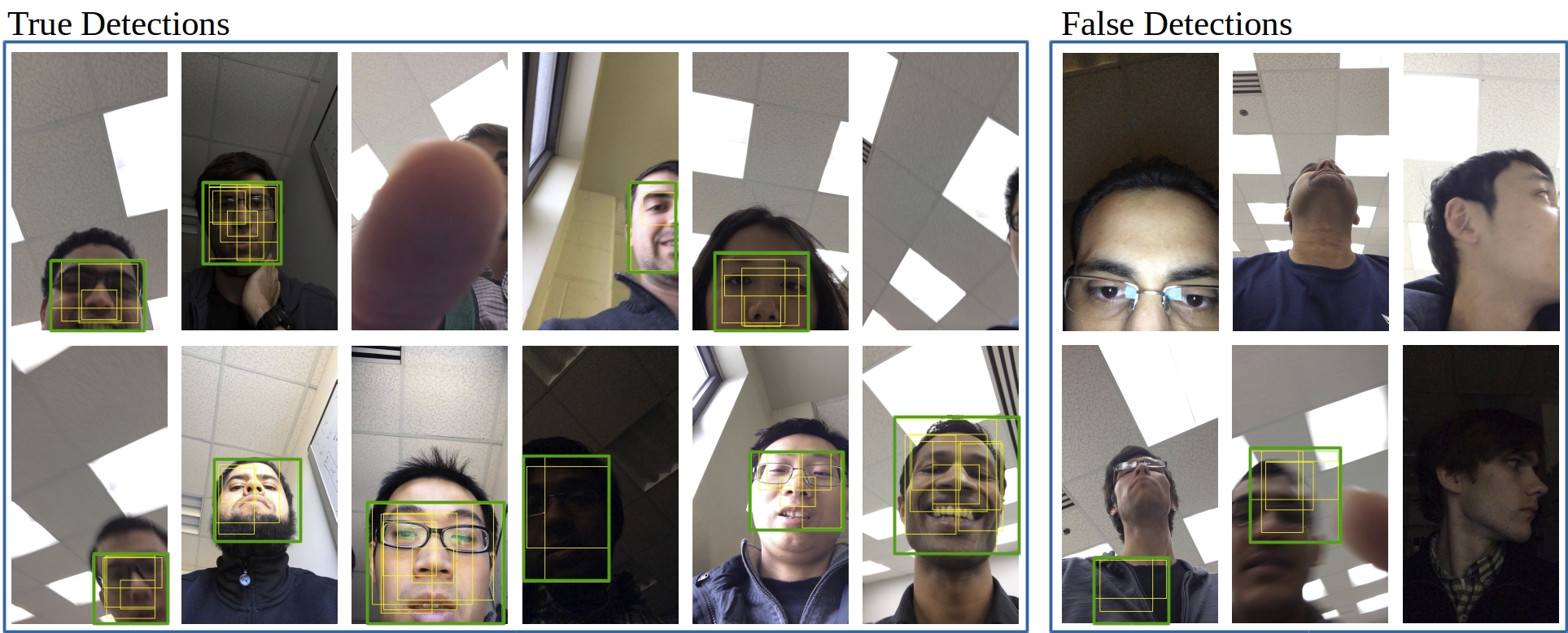}
\vskip -8pt
\caption{Detection results for $C0$. Left - true positive and true negative detection results. Yellow boxes for facial segments, green boxes for final estimated face. Right - false positive and false negative results.}
\label{DetResFigs}
\vskip -10pt
\end{figure}
\begin{figure}[t]
\centering
\includegraphics[width=0.28\textwidth]{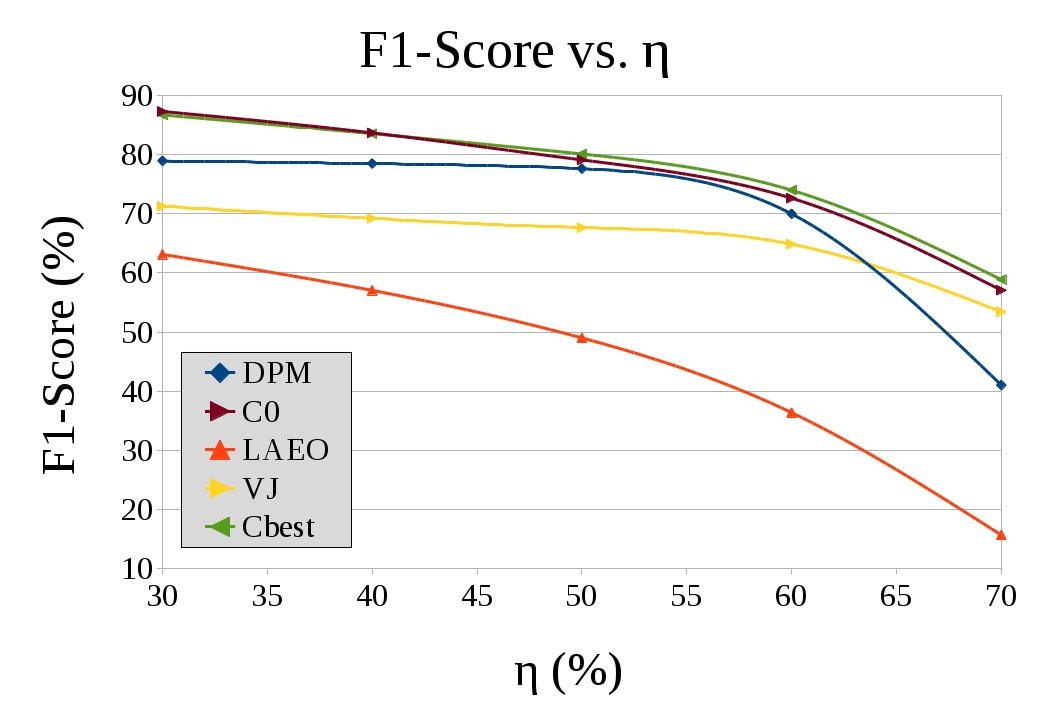}
\vskip -8pt
\caption{Comparison of F1-Score for varying overalpping threshold $\eta$ for $Cbest$ ($\zeta=20, c=2$) with state-of-the-art methods.}
\label{F1ScoreVsEta}
\vskip -10pt
\end{figure}
Some sample face detection results obtained by the proposed method for combination $C0 (\zeta=20, c=2$) are shown in Fig. \ref{DetResFigs}. In this figure, the correct detection results are depicted in the first two rows and the false detection results are shown in the bottom row.The yellow bounding boxes denote different facial segments and the green bounding box encapsulating all the yellow boxes is the final estimated face. It is apparent from this images that FSFD is robust enough to detect partial faces, such as the upper-half or left-half of a face, as well as full faces. Also, it can be seen that the operation of the proposed detector is mostly unaffected by illumination and pose variation. The images where the detector fails are mostly extreme situations where only a fragment of the face is visible, the face is critically posed, or the amount of motion blur is too much.
\begin{table}[t]
\centering
\caption{Comparison between methods at $50\%$ overlap}
\vskip -8pt
\begin{tabular}{c c c}
\hline
Method	& TPR at 	& Recall at \\
	    & $1\%$ FPR & $99\%$ Precision\\
\hline
\hline
DPM	& 0.4155	& 0.5957\\
\hline
LAEO & 0.1612		   & -		\\
\hline
FSFD $C0$ ($\zeta$ =20, c=2) &	0.4940	& 0.5577\\
\hline
FSFD $Cbest$ ($\zeta$ =20, c=2) & \textbf{0.5635}	& \textbf{0.6372}\\
\hline
\end{tabular}
\label{TPR_Recall_Compare}
\vskip -10pt
\end{table}
\begin{figure}[!htb]
\centering
\includegraphics[width=0.3\textwidth]{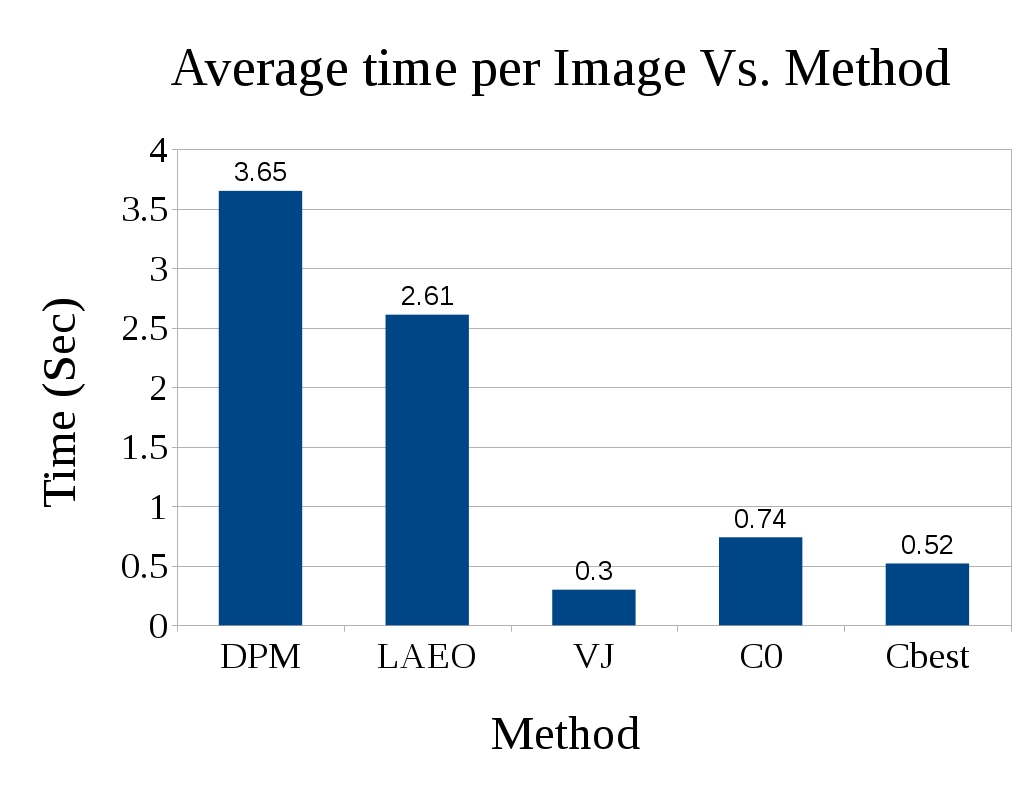}
\vskip -8pt
\caption{Time per image for different methods.}
\label{TimeVsMethod}
\vskip -10pt
\end{figure}

In Fig. \ref{F1ScoreVsEta} the F1 score of $C0$ and $Cbest$ are compared with DPM, LAEO and VJ face detection methods while varying the threshold $\eta$. It can be observed that the FSFD method outperforms all of the other methods at all values of $\eta$.  $Cbest$ especially provides superior performance at a high percentage of overlap with the ground truth.

In Table \ref{TPR_Recall_Compare}, FSFD is compared with DPM in terms of TPR at $1\%$ FPR and Recall at $99\%$ Precision. For both metrics, $Cbest$ provides significantly better performance than DPM as can be seen from the table. Note that LAEO cannot even reach $99\%$ precision with any value of recall on the dataset.
In Fig. \ref{TimeVsMethod} the average detection times per image for the four methods are compared. It can be seen that, though $Cbest$ utilizes $9$ cascade classifiers, the time consumption is not $9$-fold that of VJ which uses only one such classifier. This is because the scaling factor of VJ was set to a small value to ensure finer search in order to obtain the best results. Conversely, the scale factor of each cascade classifier in $Cbest$ was relatively big. Hence each classifier is individually weak but can produce much better outcomes when combined together. The LAEO and DPM detectors require $5$ to $7$ times more time than $Cbest$ and are not suitable for real time implementation for continuous authentication.
\section{Conclusion and Future Directions}
In this paper, a novel facial segment-based face detection technique is proposed that is suitable for face-based continuous authentication on mobile devices due to its high recall at excellent precision. A total of $14$ facial segment detectors have been trained and, an algorithm is introduced for clustering these segments to estimate a full or partially visible face. Through extensive experimentation, it is shown that the proposed method can achieve superior performance over state-of-the-art methods using fewer facial segment cascade classifiers, even as there still remains a lot of provision for speeding up the process. The future direction of this experiment is toward accurate landmark detection from overlapping facial segments and toward face authentication by fusing segment-wise verification scores.

{\small
\bibliographystyle{ieee}
\bibliography{biblio_PFD}
}
\end{document}